\documentclass[11pt,a4paper]{article}
\usepackage[hyperref]{emnlp-ijcnlp-2019}
\usepackage{times}
\usepackage{latexsym}

\usepackage{url}

\usepackage{multirow}
\usepackage{graphicx}
\usepackage{enumitem}
\usepackage{amsmath}
\usepackage{amsfonts}
\usepackage{color}
\usepackage{bm}
\usepackage{verbatim}

\usepackage[noend]{algpseudocode}
\usepackage{algorithmicx,algorithm}

\aclfinalcopy 

\title{Domain-Invariant Feature Distillation for Cross-Domain Sentiment Classification}

\author{Mengting Hu\textsuperscript{1}\thanks{\; Work performed while interning at IBM Research - China.} \quad\;\; Yike Wu\textsuperscript{1$*$} \quad\;\; Shiwan Zhao\textsuperscript{2}\thanks{\; Corresponding author.}\\ 
{\bf Honglei Guo\textsuperscript{2} \quad Renhong Cheng\textsuperscript{1} \quad Zhong Su\textsuperscript{2}} \\
\textsuperscript{1} Nankai University \quad \textsuperscript{2} IBM Research - China \\
mthu@mail.nankai.edu.cn, wuyike@dbis.nankai.edu.cn, \{zhaosw, guohl\}@cn.ibm.com \\ chengrh@nankai.edu.cn, suzhong@cn.ibm.com
}

\date{}

\begin{document}
\maketitle
\begin{abstract}

Cross-domain sentiment classification has drawn much attention in recent years. Most existing approaches focus on learning domain-invariant representations in both the source and target domains, while few of them pay attention to the domain-specific information. Despite the non-transferability of the domain-specific information, simultaneously learning domain-dependent representations can facilitate the learning of domain-invariant representations. In this paper, we focus on aspect-level cross-domain sentiment classification, and propose to distill the domain-invariant sentiment features with the help of an orthogonal domain-dependent task, i.e. aspect detection, which is built on the aspects varying widely in different domains. We conduct extensive experiments on three public datasets and the experimental results demonstrate the effectiveness of our method. 
\end{abstract}

\section{Introduction}


Sentiment classification based on deep learning methods has developed rapidly in recent years. While achieving outstanding performance, these methods always need large-scale datasets with sentiment polarity labels to train a robust sentiment classifier. However, in most cases, large-scale labeled datasets are not available in practice and manual annotation costs much. One of the solutions to this problem is \emph{cross-domain sentiment classification}, which aims to exploit the rich labeled data in one domain, i.e. \emph{source domain}, to help the sentiment analysis task in another domain lacking for or even without labeled data, i.e. \emph{target domain}.  The rationality of this solution is that the source domain and target domain share some domain-invariant knowledge which can be transferred across domains.

\begin{figure}
\centering
\includegraphics[width=0.48\textwidth]{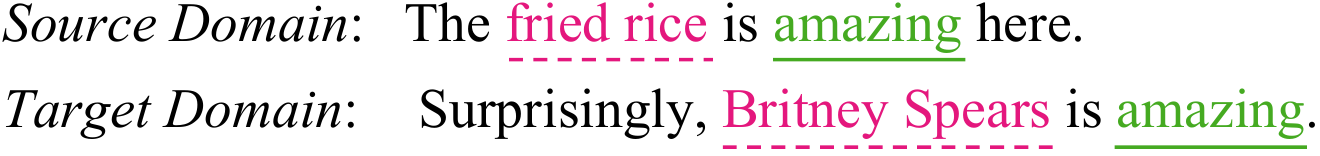}
\caption{Example sentences from the source domain (restaurant) and target domain (twitter) respectively. The sentiment expressions marked by solid lines are domain-invariant, while the aspect terms marked by dashed lines are domain-specific.}
  \label{figure-example} 
\end{figure}

Previous works on cross-domain sentiment classification mainly focus on learning the domain-invariant representations in both source and target domains, either based on manual feature selection \cite{blitzer2006domain,pan2010cross} or automatic representation learning \cite{glorot2011domain,chen2012marginalized, ganin2015unsupervised,li2017end}. The sentiment classifier, which makes decisions based on the domain-invariant features and receives the supervisory signals from the source domain, can be also applied to the target domain. We can draw an empirical conclusion: the better domain-invariant features the method obtains, the better performance it gains. However, few studies explore the usage of the domain-specific information, which is also helpful to the cross-domain sentiment classification. \newcite{peng2018cross} propose to extract the domain-invariant and domain-dependent features of the target domain data and train two classifiers accordingly, but they require a few sentiment polarity labels in the target domain, which limits the practical application of the method.



In this paper, we exploit the domain-specific information by adding an orthogonal domain-dependent task to ``distill" the domain-invariant features for cross-domain sentiment classification. The proposed method \emph{domain-invariant feature distillation} (DIFD) does not need any sentiment polarity labels in the target domain, which is more consistent with the practical settings. Specifically, we focus on the aspect-level cross-domain sentiment classification, and train a shared sentiment classifier and two respective aspect detectors in the source and target domains. We argue that the aspect detection is an orthogonal domain-dependent task with respect to the sentiment classification. As shown in Figure \ref{figure-example}, given an input sentence, the sentiment classifier predicts its sentiment polarity based on the opinion words shared by different domains, while the aspect detector identifies the aspect terms which vary significantly across domains. 
The information on which the two tasks depend are mutually exclusive in the sentence, i.e. orthogonal.
Therefore, by training these two tasks simultaneously, the aspect detectors will try to strip the domain-specific features from the input sentence and make the domain-invariant features more pure, which is helpful to the cross-domain sentiment classification. 

Moreover, we design two effective modules to boost the distillation process. One is the word-level context allocation mechanism. It modulates the importance of the words in the input sentence according to the property of different tasks. The other is the domain classifier. It tries to correctly judge which domain the domain-invariant feature comes from, while the other modules in the proposed method try to ``fool" it, and the whole framework is trained in an adversarial way. 

To summarize, the main contributions of our paper are as follows:
\begin{itemize}
    \item We distill the domain-invariant sentiment features to improve the cross-domain sentiment classification by simultaneously training an aspect detection task which striping the domain-specific aspect features from the input sentence. 
    
    \item We boost the separation process of the domain-invariant and domain-specific features by two effective modules which are the context allocation mechanism and domain classifier respectively.
    
    \item Experimental results demonstrate the effectiveness of the proposed method, and we further verify the rationality of the context allocation mechanism by visualization.
\end{itemize}


\section{Related Work}
{\bf Cross-domain sentiment analysis:} Many domain adaptation methods have been proposed for sentiment analysis. SCL \cite{blitzer2006domain} learns correspondences among features from different domains. SFA \cite{pan2010cross} aims at reducing the gap between domains by constructing a bipartite graph to model the co-occurrence relationship between domain-specific words and domain-independent words. SDA \cite{glorot2011domain} learns to extract a meaningful representation for each review in an unsupervised fashion. mSDA \cite{chen2012marginalized} is an efficient method to marginalize noise and learn features. Gradient Reversal Layer (GRL) \cite{ganin2015unsupervised,ganin2016domain,li2017end} is employed to learn domain-invariant representations by fooling the domain classifier. The replacement of gradient reversal with alternating minimization \cite{shu2018dirt} stabilizes domain adversarial training, and we employ this method as the adversarial training.

\noindent
{\bf Aspect-level sentiment domain adaptation:} To the best of our knowledge, there are two works about aspect-related cross-domain sentiment classification. \newcite{li2019exploiting} propose a method to employ abundant aspect-category data to assist the scarce aspect-term level sentiment prediction. \newcite{Zhang2019iatn} propose IATN to address that aspects have different effects in different domains. Their method predicts sentiment polarity for the whole sentence rather than a specific aspect. 

Our method concentrates on aspect-term level sentiment domain adaptation by separating the domain-specific aspect features. \newcite{bousmalis2016domain} and \newcite{liu2017adversarial} separate features into two subspaces by introducing constraints in the learned features. The difference is that our method is more fine-grained and utilizes the explicit aspect knowledge.

\begin{figure*}
\centering
\includegraphics[width=1.0\textwidth]{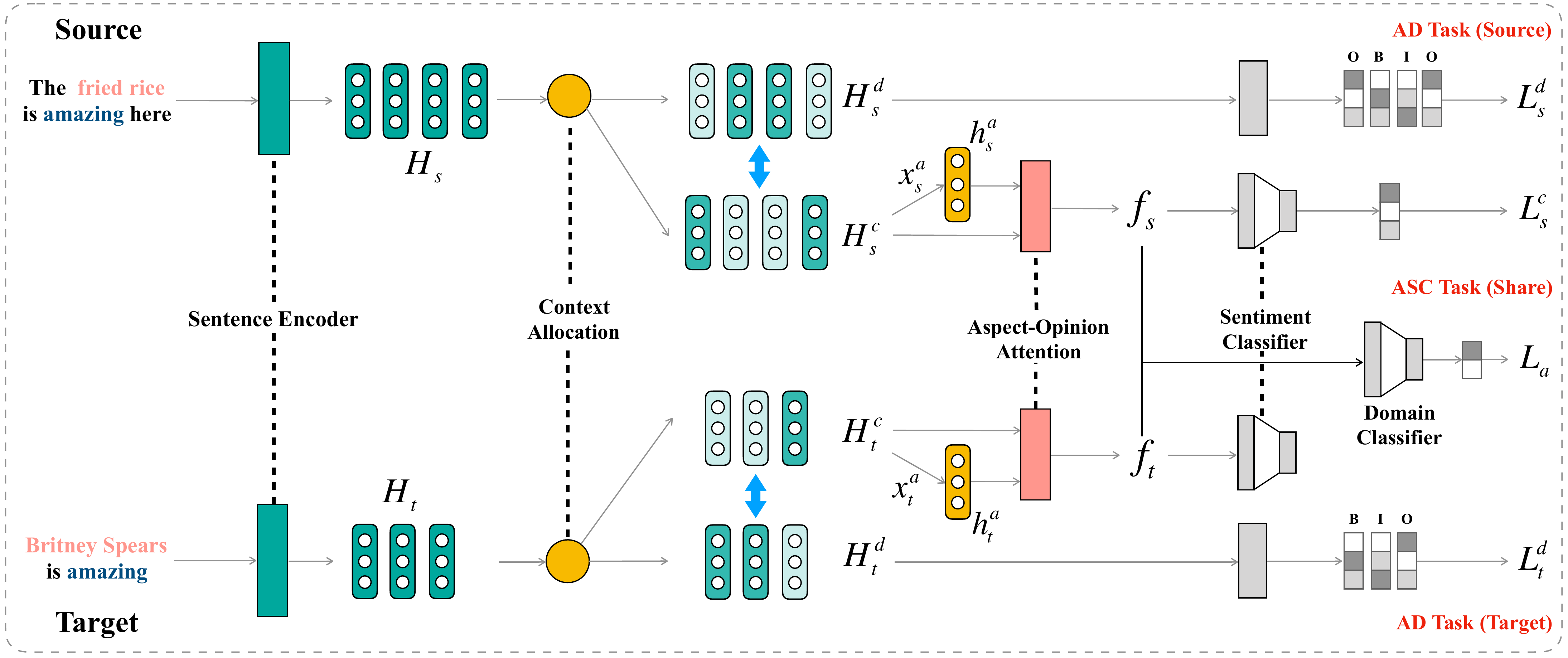}
\caption{Network Architecture. Dashed line indicates that the parameters are shared by source and target domains. The context from the  sentence encoder is divided into task-oriented contexts for ASC and AD tasks.} 
  \label{figure-network} 
\end{figure*}

\noindent
{\bf Auxiliary task for sentiment domain adaptation:} Auxiliary task has been employed to improve cross-domain sentiment analysis. \newcite{yu2016learning} use two pivot-prediction auxiliary tasks to help induce a sentence embedding, which works well across domains for sentiment classification. \newcite{yu2017leveraging} propose to jointly learn domain-independent sentence embeddings by auxiliary tasks to predict sentiment scores of domain-independent words. \newcite{chen2018cross} design auxiliary domain discriminators for better transferring knowledge between domains. These auxiliary tasks focus on directly enhancing domain-invariant features, while ours strips domain-specific features to distill domain-invariant features.

\section{Methodology}

\subsection{Formulation and Overview}
Suppose the source domain contains labeled data $\mathcal{D}_s=\{(x^k_s, a^k_s),y^k_s\}_{k=1}^{N_s}$, and the target domain contains unlabeled data $\mathcal{D}_t=\{(x^{k}_t, a^{k}_t)\}_{k=1}^{N_t}$, where $x$ is a sentence, $a$ is one of the aspects in $x$, and $y$ is the sentiment polarity label of $a$. The proposed method handles two kinds of tasks. One is the main task Aspect-level Sentiment Classification ({\bf ASC}). It learns a mapping $\mathcal{F}\colon\{x\}\to\{\bm{f}\}\to\{y\}$ shared by source and target domains, where $\bm{f}$ is the domain-invariant feature of $x$. The other is the orthogonal domain-dependent task Aspect Detection ({\bf AD}). It learns a mapping $\mathcal{G}_s\colon\{x_s\}\to\{\bm{z_s}\}\to\{a_s\}$ in the source domain, and the other one $\mathcal{G}_t\colon\{x_t\}\to\{\bm{z_t}\}\to\{a_t\}$ in the target domain, where $\bm{z}_s$ and $\bm{z}_t$ are both domain-dependent features of $x_s$ and $x_t$ respectively. The domain-invariant and domain-dependent features are orthogonal, i.e. $\bm{f}\perp\bm{z_s}$ and $\bm{f}\perp\bm{z_t}$. We facilitate the distillation of $\bm{f}$ by simultaneously learning $\mathcal{G}_s$ and $\mathcal{G}_t$ which try to strip $\bm{z_s}$ and $\bm{z_t}$ from $x$, and the purer $\bm{f}$ leads to the better $\mathcal{F}$ for the cross-domain sentiment classification.


Figure \ref{figure-network} illustrates the architecture overview of our method. Given an input sentence either from source or target domain, we first feed it into the sentence encoder to obtain its distributed representation. Then, the context allocation mechanism divides the distributed representation into two orthogonal parts: domain-invariant and domain-dependent features. Finally, the two orthogonal features are fed into their corresponding downstream tasks. Specifically, we input the domain-invariant feature into the sentiment classifier to predict the sentiment polarity, and input the domain-dependent feature into the aspect detector of the specific domain to identify the aspect terms. In addition, we add a domain classifier to the architecture. It tries to correctly judge which domain the domain-invariant feature comes from. The whole framework is trained in an adversarial way. Next we will introduce the components of our method in detail.


\subsection{Sentence Encoder}
Given an input sentence $x=\{w_1,w_2,...,w_n\}$, we first map it into an embedding sequence $\hat{E}=\{\bm{e_1},\bm{e_2},...,\bm{e_n}\}\in\mathbb{R}^{{n}\times{d_e}}$. Then we inject the positional information of each token in $x$ into $\hat{E}$ to obtain the final embedded representation $E$, following the Position Encoding (PE) method in the work \cite{vaswani2017attention}:
\begin{equation}
    \begin{split}
        PE(pos,2i)&=sin(pos/10000^{2i/d_e}) \\
        PE(pos,2i+1)&=cos(pos/10000^{2i/d_e}) \\
        E&=\hat{E}+PE\\
    \end{split}
\end{equation}
where $pos$ is the word position in the sentence and $i$ is the $i$-th dimension of $d_e$. We consider that the injected positional information can facilitate the aspect-level sentiment classification, based on the observation that sentiment words tend to be close to its related aspect terms \cite{tang2016aspect, chen2017recurrent}.

Next we employ a Bi-directional LSTM (BiLSTM) \cite{Graves2013Speech} to encode $E$ into the contextualized sequence representation $H=[\bm{h_1},\bm{h_2},...\bm{h_n}]\in\mathbb{R}^{n\times{2d_h}}$, which preserves the contextual information of each token in the input sentence.

We unify the embedding layer and BiLSTM as the sentence encoder in which different tasks or domains all share the same weights. The advantages of sharing the weights are two-fold: first, different tasks in the same domain can benefit from each other in a multi-task manner; second, distilling the domain-invariant feature from a common transformation is more simple.

\subsection{Context Allocation (CA)}
In an input sentence, some words have a strong bias towards domain-specific information, such as the aspect terms, e.g. \emph{``pizza"} in Restaurant domain, while others focus on the domain-invariant knowledge, such as the opinion words, e.g. \emph{``amazing"}. Meanwhile, the ASC task and AD task exactly require orthogonal information as discussed before. Therefore, we argue that different words contribute differently according to the property of the downstream task. To facilitate the distillation of the domain invariant features, we propose a Context Allocation (CA) mechanism to allocate different weights on the same word in different downstream tasks. The values of the weights depend on how the information contained in the word matches the need of the specific task. Concretely, at each time step $i$, the module divides the contextualized representation $h_i$ of word $w_i$ into the sentiment-dominant context $h_i^c$ and aspect-dominant context $h_i^d$ as follows:
\begin{gather}
    \bm{h_i^c} = \beta_i^c \bm{h_i},\\
    \bm{h_i^d} = \beta_i^d \bm{h_i}, \\
    \beta_i^c + \beta_i^d = 1,\quad i \in \{1,2,...,n\}.
\end{gather}
The two-dimensional vector $\beta_i = (\beta_i^c, \beta_i^d)$ is normalized considering that the domain-specific information and domain-invariant knowledge are mutually exclusive. It reflects the importance of $w_i$ on the ASC task and AD task respectively, and is calculated on $h_i$ as follows:
\begin{equation}
\setlength{\abovedisplayskip}{4pt}  
\setlength{\belowdisplayskip}{4pt}  
    \beta_i = \mathrm{softmax}(W_b\mathrm{tanh}(W_a\bm{h_i}^\mathrm{T})),
\end{equation}
where $W_a\in\mathbb{R}^{2d_h\times{2d_h}}$ and $W_b\in\mathbb{R}^{2\times{2d_h}}$. The whole division process at all time steps can be formulated in the following form:
\begin{gather}
     \left[\begin{matrix}
    H_c\\
    H_d
    \end{matrix}\right]
     =  \boldsymbol{\beta} \cdot 
     \left[\begin{matrix}
    H \\
    H
    \end{matrix}\right],
\end{gather}
where $\boldsymbol{\beta}=[\beta_1,\beta_2,...,\beta_n]$. The sentiment-dominant context $H^c\in\mathbb{R}^{n\times{2d_h}}$ and aspect-dominant context $H^d\in\mathbb{R}^{n\times{2d_h}}$ of the input sentence are then fed into the ASC task and AD task for downstream processing respectively.






\subsection{Aspect-level Sentiment Classification (ASC) Task}
{\bf Aspect-Opinion Attention} In the ASC task, we design an attention mechanism to model the relationship between the position of the aspect terms and their corresponding opinion words. For a specific aspect term, the domain-invariant feature based on the aspect-opinion attention contains more information of its corresponding opinion words, which is beneficial to the final aspect-level sentiment classification. Specifically, we first calculate the position representation of a specific aspect term with its position $x^a$ and the sentiment-dominant context $H^c$:
\begin{equation}
    \boldsymbol{h^a} = H^cx^a
\end{equation}

where $x^a=\{0_1,...,1_{i+1},...,1_{i+m},...0_n\}$ represents the word positions of an aspect sub-sequence in the input sentence $x$ with non-zero values and $m$ is the length of the aspect. Then the representation $\boldsymbol{h^a}\in\mathbb{R}^{2d_h}$ is further utilized to calculate the sentiment-dominant feature $\boldsymbol{f}$, which is domain-invariant and should be aligned across source and target domains. 

\begin{equation}
\setlength{\abovedisplayskip}{4pt}  
\setlength{\belowdisplayskip}{4pt}  
    \begin{split}
    \gamma_i &= \mathrm{tanh}(\bm{h_i^c}\cdot{W_p}\cdot{\boldsymbol{h^a}}+b_p) \\
    \boldsymbol{f} &= \sum_{i=1}^n{\gamma_i\bm{h_i^c}}
    \end{split}
\end{equation}

where $\gamma_i$ reflects how much the word $w_i$ corresponds with the opinion on the aspect term, and $W_p\in\mathbb{R}^{2d_h\times{2d_h}}$ and $b_p\in\mathbb{R}^1$ are weight matrix and bias respectively. 

{\bf Sentiment Classification Loss}  The sentiment-dominant features $\boldsymbol{f_s}$ and $\boldsymbol{f_t}$ generated from the source and target domains respectively share the same sentiment classifier. Note that the source domain data has sentiment polarity label, while the target domain is unlabeled. Thus we train the sentiment classifier only with the labeled data in the source domain, while utilize it for inference in both source and target domains. The training objective of the sentiment classifier is to minimize the following loss on the source domain dataset, which is marked as $L_s^c$:

\begin{equation}
\setlength{\abovedisplayskip}{4pt}  
\setlength{\belowdisplayskip}{4pt}  
    L_s^c = -\frac{1}{N_s}\sum^{N_s}\mathrm{log}{P{(y|\boldsymbol{f_s})}}
\end{equation}

where $y$ is the ground-truth sentiment polarity label. For simplicity, we omit the enumerate number of the instance in the loss equation.


{\bf Domain Adversarial Loss} The domain classifier maps the sentiment-dominant feature $\boldsymbol{f}$ into a two-dimensional normalized value $\boldsymbol{y} = (y_s, y_t)$, which indicates the probability that $\boldsymbol{f}$ comes from the source and target domains respectively. The ground-truth domain label is $\boldsymbol{g_s} = (1,0)$ for instances in the source domain, and $\boldsymbol{g_t} = (0,1)$ in the target domain. The training objective of the domain classifier is to minimize the following loss on both source and target domain datasets, which is marked as $L_a^{\theta_D}$:
\begin{equation}
     L_a^{\theta_D} = -\frac{1}{N_s}\sum^{N_s}\boldsymbol{g_s}{\mathrm{log}}{\boldsymbol{y}} -\frac{1}{N_t}\sum^{N_t}\boldsymbol{g_t}{\mathrm{log}}{\boldsymbol{y}}.
    \label{dc_loss}
\end{equation}

The part in our architecture which joins the generating process of $\boldsymbol{f}$ (including Sentence Encoder, Context Allocation and Aspect-Opinion Allocation in Figure \ref{figure-network}) can be regarded as a domain-invariant feature extractor, which works with the domain classifier in an adversarial way. To further accelerate the distillation process of the domain-invariant features, we also introduce an adversarial loss of the domain classifier for the feature extractor. Specifically, we calculate the loss in Equation \ref{dc_loss} with the flipped domain labels inspired by the work \cite{shu2018dirt}: 

\begin{equation}
    L_a^{\theta_F} = -\frac{1}{N_s}\sum^{N_s}\boldsymbol{g_t}{\mathrm{log}}{\boldsymbol{y}} -\frac{1}{N_t}\sum^{N_t}\boldsymbol{g_s}{\mathrm{log}}{\boldsymbol{y}}.
\end{equation}



\begin{algorithm}[t]
	\caption{Adversarial Training}
	\hspace*{0.02in} {\bf Input:}
	labeled $\mathcal{D}_s$ and unlabeled $\mathcal{D}_t$
	\begin{algorithmic}[1]
		\Repeat
		\State {\bf Train} All parameters except Domain Classifier with $L$;
		\State {\bf Train} Domain Classifier with $\lambda^aL_a^{\theta_D}$;
		\Until \emph{performance on the validation set does not improve in 10 epochs.}
	\end{algorithmic}
\end{algorithm}

\subsection{Aspect Detection (AD) Task}
We model the AD task as a sequence labelling problem, and each word in the sentence is marked as a tag in $\{\mathrm{B,I,O}\}$, which means the word is at the beginning ($\mathrm{B}$) or the inside ($\mathrm{I}$) of an aspect term or other word ($\mathrm{O}$). In this way, we can detect all the aspect terms of an input sentence in one forward pass. Specifically, we first linearly transform the aspect-dominant hidden state $\bm{h^d}$ into a three-dimensional vector. Then we calculate the aspect detection loss of the source domain as follows: 

\begin{equation}
L^d_s=-\frac{1}{N_s}{\sum^{N_s}}\frac{1}{n}{\sum^n}{\lambda_l}\mathrm{log}P(y^d|\bm{h^d})
\end{equation}

where $y^d$ is the ground-truth aspect label, $n$ is the sentence length and $\lambda_l$ is the weight of different labels. The weight $\lambda_l$ aims to solve the class imbalance problem because the words labeled by $\mathrm{O}$ usually make up the majority of one sentence. It is dynamically calculated in the training phase according to the ratio of the words with a specific label in each batch. Henceforth we denote the loss of the AD task in the target domain as $L^d_t$. 


\subsection{Training}
We combine each component loss into an overall object function:
\begin{equation}
    L = L_s^c + \lambda^aL_a^{\theta_F} + \lambda^d(L^d_s + L^d_t)
\end{equation}

where $\lambda^a$ and $\lambda^d$ balance the effect of the domain classifier and the auxiliary task (i.e. aspect detection). $L$ and $\lambda^aL_a^{\theta_D}$ are alternatively optimized. The aspect-level sentiment analysis in the unlabeled target domain is predicted by the ASC task.

\section{Experiments}

\subsection{Datasets}
To make an extensive evaluation, we employ three different datasets: Restaurants (R) and Laptops (L) from SemEval 2014 task 4 \cite{Pontiki2014SemEval}, and Twitters (T) from the work \cite{dong2014adaptive}. The statistics of these three datasets are shown in Table \ref{table-dataset}. Specifically, we collect the aspect-term level sentences and corresponding labels from these datasets. Comparing aspect terms in these three datasets, we find more than 98\% aspect terms are different between Restaurants and Laptops domains, and there exists no same aspect between Restaurants and Twitters, also only 0.09\% same aspects between Laptops and Twitters. This indicates that the aspect terms vary violently in different domains.

\begin{table}[t!]
\begin{center}
\setlength{\tabcolsep}{1.0mm}{
\begin{tabular} {|c|c|ccc|c|}
\hline
    \multicolumn{2}{|c|}{Dataset} & \#Pos & \#Neg & \#Neu & Total \\ 
	    \hline
	    \multirow{2}{*}{Restaurants (R)} & Train & 2164 & 805 & 633 & 3502 \\
	                                 & Test  & 728  & 196 & 196 & 1120 \\
	    \hline
	    \multirow{2}{*}{Laptops (L)}     & Train & 987  & 866 & 460 & 2313 \\
	                                 & Test  & 341  & 128 & 169 & 638  \\
	    \hline
	    \multirow{2}{*}{Twitters (T)}    & Train & 1561 & 1560& 3127& 6248 \\
	                                 & Test  & 173  & 173 & 346 & 692  \\
	\hline
\end{tabular}}
\end{center}
\caption{\label{table-dataset} Datasets statistics. \#Pos, \#Neg, \#Neu represent the number of instances with positive, negative and neutral polarities.}
\end{table}

\begin{table*}[t!]
\begin{center}
\setlength{\tabcolsep}{1.4mm}{
\begin{tabular} {|c|cc|cc|cc|cc|cc|cc|}
\hline
    \multirow{2}{*}{Models} & \multicolumn{2}{c|}{R$\rightarrow$L} & \multicolumn{2}{c|}{L$\rightarrow$R} & \multicolumn{2}{c|}{R$\rightarrow$T} & \multicolumn{2}{c|}{T$\rightarrow$R} & \multicolumn{2}{c|}{L$\rightarrow$T} & \multicolumn{2}{c|}{T$\rightarrow$L} \\  
    \cline{2-13}
    & Acc & F1 & Acc & F1 & Acc & F1 & Acc & F1 & Acc & F1 & Acc & F1 \\
    \hline
    ATAE-LSTM & 56.56 & 47.71 & 63.21 & 46.01 & 34.97 & 33.82 & 50.78 & 44.06 & 40.43 & 39.82 & 42.39 & 40.73 \\
    MemNet & 57.17 & 47.79 & 65.97 & 49.28 & 34.47 & 32.37 & 50.36 & 42.68 & 44.94 & 44.37 & 36.36 & 34.96 \\
    RAM & 58.32 & 43.98 & 54.68 & 26.81 & 33.18 & 29.96 & 48.18 & 43.23 & 44.87 & 44.06 & 42.73 & 42.51 \\
    GACE & 61.74 & 50.39 & 66.60 & 49.07 & 32.87 & 29.34 & 45.62 & 37.52 & 44.93 & 45.18 & 47.98 & 42.39 \\
    AT-LSTM & 60.62 & 45.38 & 66.75 & 46.99 & 32.98 & 28.98 & 50.64 & 43.64 & 38.47 & 36.78 & 47.34 & 42.87 \\
    IAN & 60.39 & 50.69 & 66.50 & 47.97 & 35.09 & 33.02 & 51.14 & 44.46 & 44.16 & 44.34 & 43.88 & 40.16 \\
    \hline
    IATN & 62.05 & 46.37 & 67.13 & 51.64 & 34.14 & 30.55 & 40.13 & 37.79 & 44.80 & 44.69 & 46.12 & 42.91 \\
    \hline
    DIFD(S) & 63.81 & 57.74 & 68.30 & 53.66 & 36.89 & 33.74 & 55.63 & 44.10 & 45.73 & 46.17 & 44.53 & 43.76 \\
    DIFD & {\bf 64.86} & {\bf 60.51} & {\bf 68.53} & {\bf 57.31} & {\bf 40.13} & {\bf 38.85} & {\bf 57.60} & {\bf 46.59} & {\bf 47.32} & {\bf 47.31} & {\bf 48.97} & {\bf 47.56} \\
\hline
\end{tabular}}
\end{center}
\caption{\label{table-result} Evaluation results of baselines in terms of accuracy (\%) and macro-f1 (\%).}
\end{table*}

\begin{table*}[t!]
\begin{center}
\setlength{\tabcolsep}{1.0mm}{
\begin{tabular} {|c|cc|cc|cc|cc|cc|cc|}
\hline
    \multirow{2}{*}{Models} & \multicolumn{2}{c|}{R$\rightarrow$L} & \multicolumn{2}{c|}{L$\rightarrow$R} & \multicolumn{2}{c|}{R$\rightarrow$T} & \multicolumn{2}{c|}{T$\rightarrow$R} & \multicolumn{2}{c|}{L$\rightarrow$T} & \multicolumn{2}{c|}{T$\rightarrow$L} \\
    \cline{2-13}
    & Acc & F1 & Acc & F1 & Acc & F1 & Acc & F1 & Acc & F1 & Acc & F1 \\
    \hline
    ASC+AT & 62.86 & 56.47 & 67.66 & 49.84 & 35.86 & 33.43 & 52.33 & 44.41 & 44.15 & 44.31 & 42.46 & 37.94 \\
    DIFD-CA & 64.18 & 57.59 & 68.17 & 53.42 & 34.27 & 31.52 & 44.39 & 41.79 & 44.14 & 44.25 & 43.54 & 43.60 \\
    DIFD-AT & 63.47 & 60.09 & 68.15 & 57.27 & 40.76 & 38.66 & {\bf 59.76} & 45.08 & 45.73 & 45.91 & 47.68 & 40.20 \\
    \hline
    DIFD-AT+MMD & 64.25 & 58.79 & 63.13 & 56.49 & 39.44 & 38.69 & 46.93 & 44.44 & {\bf 47.81} & {\bf 47.91} & 42.87 & 38.69 \\
    DIFD-AT+CORAL & 63.77 & 58.61 & 68.30 & 53.46 & {\bf 44.65} & {\bf 42.83} & 57.54 & 46.41 & 46.96 & 46.96 & 38.80 & 35.97 \\
    \hline
    DIFD & {\bf 64.86} & {\bf 60.51} & {\bf 68.53} & {\bf 57.31} & 40.13 & 38.85 & 57.60 & {\bf 46.59} & 47.32 & 47.31 & {\bf 48.97} & {\bf 47.56} \\
\hline
\end{tabular}}
\end{center}
\caption{\label{table-result-var} Evaluation results of variants of our model in terms of accuracy(\%) and macro-f1(\%). The minus sign (-) means to remove the module, and the addition (+) means to add the module.}
\end{table*}

\subsection{Experimental Settings}
To evaluate our proposed method, we construct six aspect-level sentiment transfer tasks: R$\rightarrow$L, L$\rightarrow$R, R$\rightarrow$T, T$\rightarrow$R, L$\rightarrow$T, T$\rightarrow$L. The arrow indicates the transfer direction from the source domain to the target domain. For each transfer pair $\mathcal{D}_s\rightarrow{\mathcal{D}_t}$, the training set is composed of two parts: one is the labeled training set in $\mathcal{D}_s$, and the other is all unlabeled data which only contain the aspect term information in $\mathcal{D}_t$. The test set in $\mathcal{D}_s$ is employed as the validation set. The reported results are evaluated on all the data of $\mathcal{D}_t$.

The word embeddings are initialized with 100-dimension Glove vectors \cite{pennington2014glove} and fine-tuned during the training. The model hidden size $d_h$ is set to be 64. The model is optimized by the SGD method with the learning rate of 0.01. The batch size is 32. We employ ReLU as the activation function.

We adopt an early stop strategy during training if the performance on the validation set does not improve in 10 epochs, and the best model is choosed for evaluation.

\subsection{Compared Methods}
We compare with extensive baselines to validate the effectiveness of the proposed method. Some variants of our approach are also compared for analyzing the impacts of individual components.

\noindent
{\bf Transfer Baseline:}
The aspect-level cross-domain sentiment classification has been rarely explored. We choose the state-of-the-art method IATN \cite{Zhang2019iatn} which has the most similar settings with our method as the transfer baseline. It proposes to incorporate the information of both sentences and aspect terms in the cross-domain sentiment classification.

\noindent
{\bf Non-Transfer Baselines:}
The non-transfer baselines are all representative methods in recent years for the aspect-level sentiment classification in a single domain. We train the models on the training set of the source domain, and directly test them in the target domain without domain adaptation.
\begin{itemize}
    \vspace{-6pt}
    \item {\bf AT-LSTM} \cite{wang2016attention}: It utilizes the attention mechanism to generate an aspect-specific sentence representation. 
    
    \vspace{-6pt}
    \item {\bf ATAE-LSTM} \cite{wang2016attention}: It also employs attention. The difference with AT-LSTM is that the aspect embedding is as input to LSTM.
    
    \vspace{-6pt}
    \item {\bf MemNet} \cite{tang2016aspect}: It employs a memory network with multi-hops attentions and predicts the sentiment based on the top-most context representations.
    
    \vspace{-6pt}
    \item {\bf IAN} \cite{ma2017interactive}: It adopts interactive attention mechanism to learn the representations of the context and the aspect respectively.
    
    \vspace{-6pt}
    \item {\bf RAM} \cite{chen2017recurrent}: It employs multiple attentions with a GRU cell to non-linearly combine the aggregation of word features in each layer.
    
    \vspace{-6pt}
    \item {\bf GACE} \cite{xue2018aspect}: It is based on the convolutional neural network with gating mechanisms.

\end{itemize}

{\bf Variants of Our Method:}
\begin{itemize}
    
    \vspace{-6pt}
    \item {\bf ASC+AT} (ASC with adversarial training): A single task which handles the ASC task with adversarial training.
    
    \vspace{-6pt}
    \item {\bf DIFD:} The proposed method in this work. 
    
    \vspace{-6pt}
    \item {\bf DIFD(S)}: It contains components of the source domain from DIFD and is trained only by the source domain data.
    
    \vspace{-6pt}
    \item {\bf DIFD-CA}: DIFD without context allocation.
    
    \vspace{-6pt}
    \item {\bf DIFD-AT}: DIFD without adversarial training.
    
    \vspace{-6pt}
    \item {\bf DIFD-AT+MMD}: Replace the adversarial training with Maximum Mean Discrepancy (MMD) \cite{tzeng2014deep}.
    
    \vspace{-6pt}
    \item {\bf DIFD-AT+CORAL}: Replace the adversarial training with CORAL  \cite{sun2016return}.

\end{itemize}

\subsection{Experimental Analysis}
We report the classification accuracy and macro-f1 of various methods in Table \ref{table-result} and Table \ref{table-result-var}, and the best scores on each metric are marked in bold. 
To validate the effectiveness of our method, we analyze the results from the following perspectives.

\noindent
{\bf Compare with the baselines:} 
We display the comparison results with baselines in Table \ref{table-result}. Comparing with the transfer baseline IATN, we observe that DIFD significantly outperforms IATN on all metrics by +5.51\% accuracy and +7.36\% macro-f1 on average. This shows that the distillation of domain-invariant features really facilitates the transfer of sentiment information across domains. In addition, for fair comparison with the non-transfer methods which only exploits the source domain data, we also train our DIFD model without the target domain data and denote this variant as DIFD(S). We observe that DIFD(S) outperforms all the non-transfer baselines on most metrics. It is worth noting that, compared to a strong baseline IAN, DIFD(S) achieve significant improvement by +4.49\% accuracy on T$\rightarrow$R and +7.05\% macro-f1 on R$\rightarrow$L. This verifies that the orthogonal task is helpful in striping the domain-specific features from the source domain and effective for accelerating the domain adaptation.


\begin{figure*}
\centering
\includegraphics[width=1.0\textwidth]{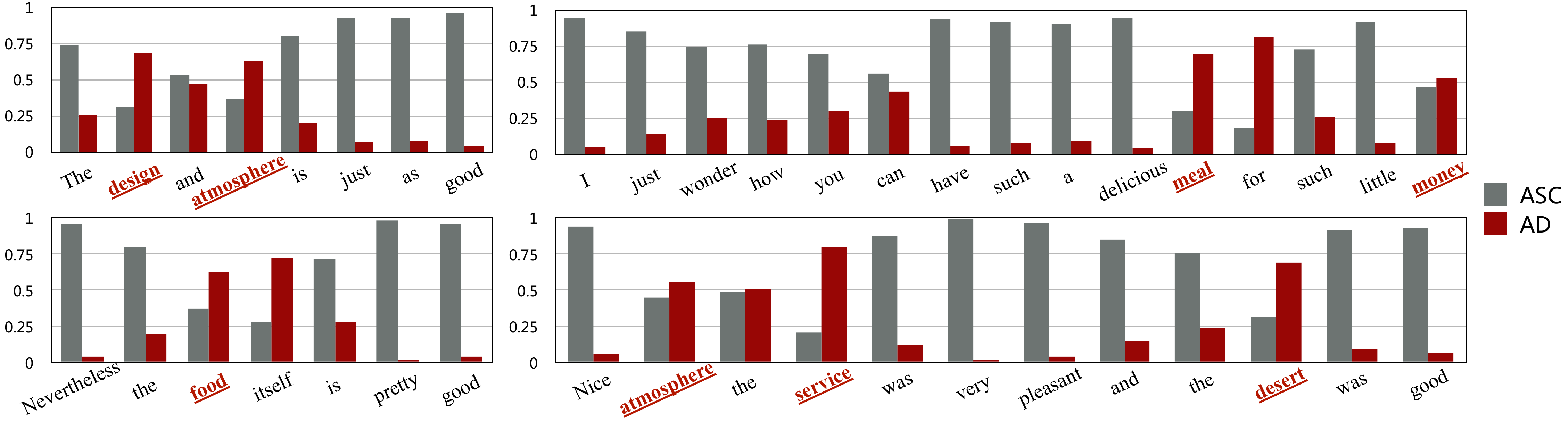}
\caption{Visualization of context allocation weights of DIFD for different tokens within a sequence.} 
  \label{figure-case} 
\end{figure*}

\noindent
{\bf Compare the variants of our method:} 
The results of the variants of our method  are reported in Table \ref{table-result-var}. We first observe that DIFD outperforms ASC+AT on all metrics significantly. This validates that the orthogonality really helps to distill the domain-invariant features and improve the performance of the cross-domain sentiment classification. 

Then we can see that DIFD-CA performs much worse than DIFD, which reveals that the context allocation mechanism plays an important role in our method. We further visualize the allocation scores in Figure \ref{figure-case} and the result also indicates that the reasonability of the CA module. The gray tokens and red tokens have a bias towards ASC task and AD task respectively. The allocation scores are consistent with the bias of words: red tokens get larger scores for the aspect detection task, while gray tokens get larger scores for opinion expressions. This shows that our model generates task-oriented contexts successfully.

Finally, DIFD also achieves improvement over DIFD-AT on most metrics. This indicates that adversarial training with the domain classifier promotes the distillation process of the domain-invariant features. To further validate the effectiveness of adversarial training, we also try to directly minimize the divergence between domain-invariant features from source and target domains based on MMD and CORAL. Comparing with DIFD-AT+MMD and DIFD-AT+CORAL, DIFD is more robust considering that DIFD outperforms the two methods on most experimental settings. 





\begin{figure}
\setlength{\belowcaptionskip}{-0.1cm}   
\centering
\includegraphics[width=0.45\textwidth]{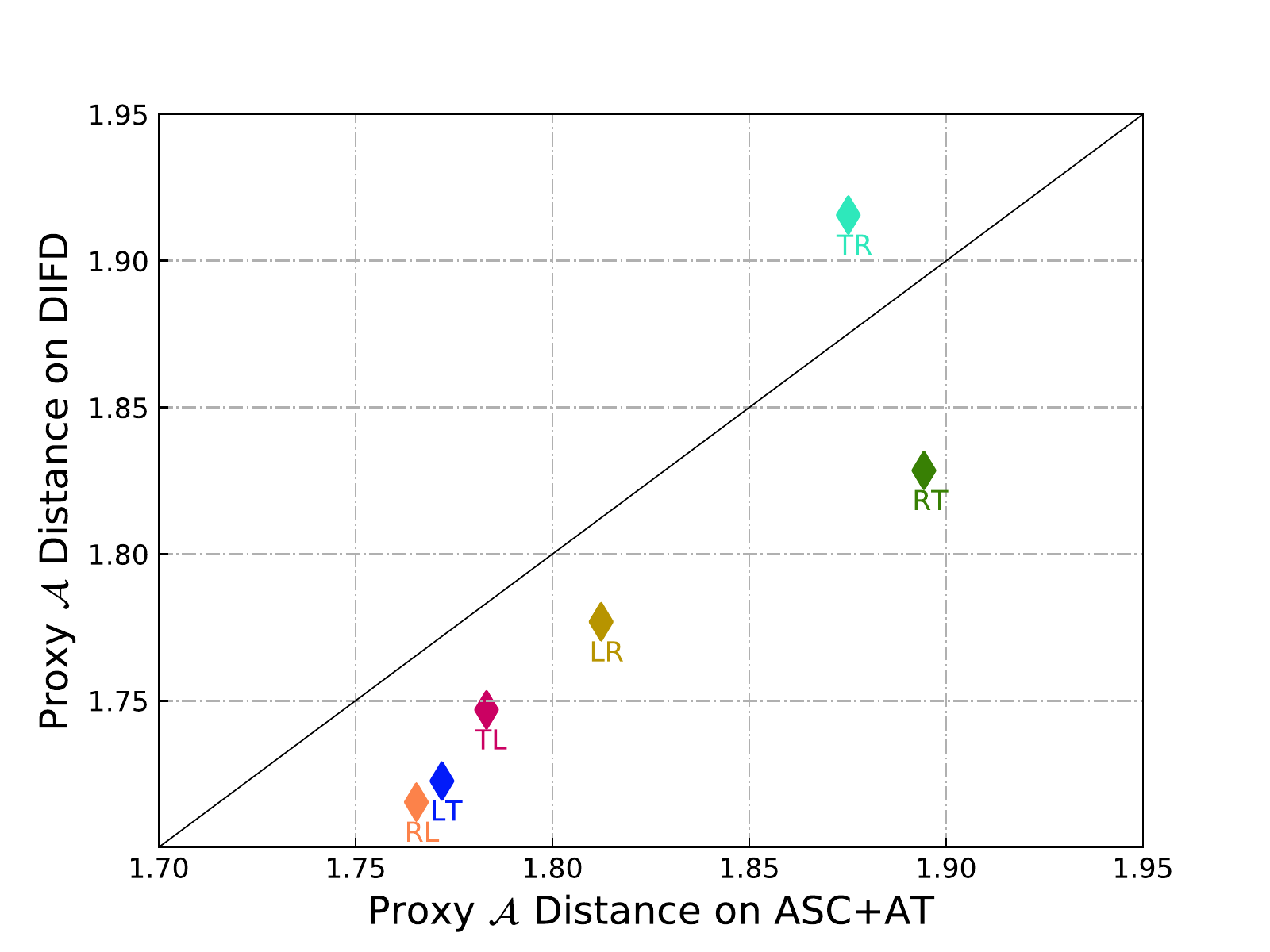}
\caption{Proxy $\mathcal{A}$ Distance between domain-invariant features for the 6 different pairs.}
  \label{figure-distance-f} 
\end{figure}



\subsection{Transfer Distance Analysis}
\begin{figure}
\setlength{\belowcaptionskip}{-0.1cm}   
\centering
\includegraphics[width=0.45\textwidth]{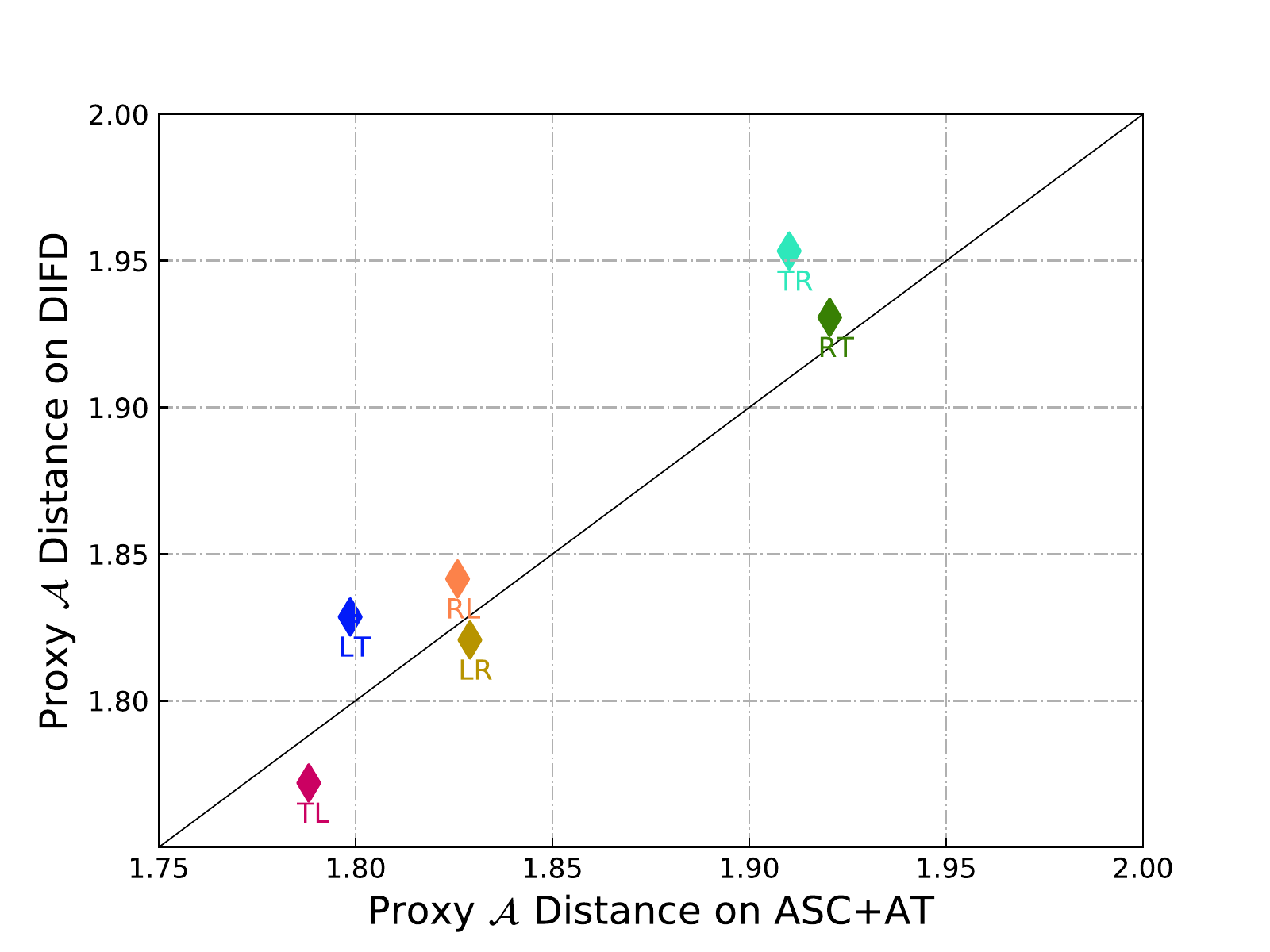}
\caption{Proxy $\mathcal{A}$ Distance between domain-specific features for the 6 different pairs.}
  \label{figure-distance-ad} 
\end{figure}

In this section, we analyze the similarity of features between domains. We exploit the $\mathcal{A}$-distance \cite{ben2007analysis} to measure the similarity between two probability distributions. The proxy $\mathcal{A}$-distance is 2(1-2$\epsilon$), where $\epsilon$ is the generalization error of a classifier (a linear SVM) trained on the binary classification problem to distinguish inputs between the two domains. We focus on the methods ASC+AT and DIFD, and first compare the similarity of domain-invariant features $\boldsymbol{f_s}$ and $\boldsymbol{f_t}$. Figure \ref{figure-distance-f} reports the results for each pair of domains. The proxy $\mathcal{A}$-distance on DIFD is generally smaller than its corresponding value on ASC+AT. This indicates that DIFD can learn purer domain-invariant features than ASC+AT. Secondly, we compare the domain-specific features learned by ASC+AT and DIFD, which are represented by the average hidden state of BiLSTM in ASC+AT and the average aspect-context $H^d$ in DIFD respectively. Figure \ref{figure-distance-ad} reports the results for each pair of domains. The proxy $\mathcal{A}$-distance on DIFD is generally larger than its corresponding value on ASC+AT, which demonstrates that DIFD can strip more domain-specific information by the aspect detection task than ASC+AT. 

There are exceptions in both Figures, i.e., TR in Figure \ref{figure-distance-f}, TL and LR in Figure \ref{figure-distance-ad}. A possible explanation is that the balance between ASC and AD losses causes some domain-specific information to remain in the domain-invariant space, and vice versa.

\section{Conclusion}
In this work, we study the problem of aspect-level cross-domain sentiment analysis and propose a domain-invariant feature distillation method which simultaneously learns domain-invariant and domain-specific features. With the help of the orthogonal domain-dependent task (i.e., aspect detection), the aspect sentiment classification task can learn better domain-invariant features and improve transfer performance. Experimental results clearly verify the effectiveness of our method.

\section{Acknowledgement}
This work is supported by National Science and Technology Major Project, China (Grant No. 2018YFB0204304).


\bibliography{emnlp-ijcnlp-2019}
\bibliographystyle{acl_natbib}

\end{document}